\documentclass[10pt,twocolumn,letterpaper]{article}

\usepackage{cvpr}              

\usepackage{graphicx}
\usepackage{amsmath}
\usepackage{amssymb}
\usepackage{booktabs}
\usepackage{float}
\graphicspath{ {./images/} }
\usepackage[pagebackref,breaklinks,colorlinks]{hyperref}

\usepackage[capitalize]{cleveref}
\crefname{section}{Sec.}{Secs.}
\Crefname{section}{Section}{Sections}
\Crefname{table}{Table}{Tables}
\crefname{table}{Tab.}{Tabs.}

\def\cvprPaperID{28} 
\def\confName{ICCV}
\def\confYear{2023}

\begin{document}

\title{SATHUR: Self Augmenting Task Hallucinal Unified Representation\\ for Generalized Class Incremental Learning}



\author{Sathursan Kanagarajah \hspace{1cm} Thanuja Ambegoda \hspace{1cm} Ranga Rodrigo\\
University of Moratuwa, Sri Lanka\\
{\tt\small ksathursan1408@gmail.com, thanujaa@uom.lk, ranga@uom.lk}
}

\maketitle

\begin{abstract}
Class Incremental Learning (CIL) is inspired by the human ability to learn new classes without forgetting previous ones. CIL becomes more challenging in real-world scenarios when the samples in each incremental step are imbalanced. This creates another branch of problem, called Generalized Class Incremental Learning (GCIL) where each incremental step is structured more realistically.
Grow When Required (GWR) network, a type of Self-Organizing Map (SOM), dynamically create and remove nodes and edges for adaptive learning. GWR performs incremental learning from feature vectors extracted by a Convolutional Neural Network (CNN), which acts as a feature extractor. The inherent ability of GWR to form distinct clusters, each corresponding to a class in the feature vector space, regardless of the order of samples or class imbalances, is well suited to achieving GCIL.
To enhance GWR's classification performance, a high-quality feature extractor is required. However, when the convolutional layers are adapted at each incremental step, the GWR nodes corresponding to prior knowledge are subject to near-invalidation.
This work introduces the \emph{Self Augmenting Task Hallucinal Unified Representation (SATHUR)}, which re-initializes the GWR network at each incremental step, aligning it with the current feature extractor. Comprehensive experimental results demonstrate that our proposed method significantly outperforms other state-of-the-art GCIL methods on CIFAR-100 and CORe50 datasets.\\
\end{abstract}
\vspace{-0.9cm}
\section{Introduction}
\label{sec:intro}
Humans and animals have an astonishing ability to continually acquire, process, and update knowledge throughout their lifetime. The ability to continually learn over time by accumulating new knowledge while retaining and utilizing previously learned knowledge is referred to as Incremental Learning (IL)~\cite{chen2018lifelong}. Convolutional Neural Network (CNN) has achieved expert-level performances in various computer vision problems. In some challenges like image classification, their performance has even surpassed that of humans. However, when CNN models attempt to learn tasks incrementally, they partially or completely forget the previously learned knowledge, a phenomenon termed catastrophic forgetting~\cite{mccloskey1989catastrophic,mcrae1993catastrophic,french1999catastrophic}.
\begin{figure}[t]
\begin{center}
   \includegraphics[width=0.47\textwidth]{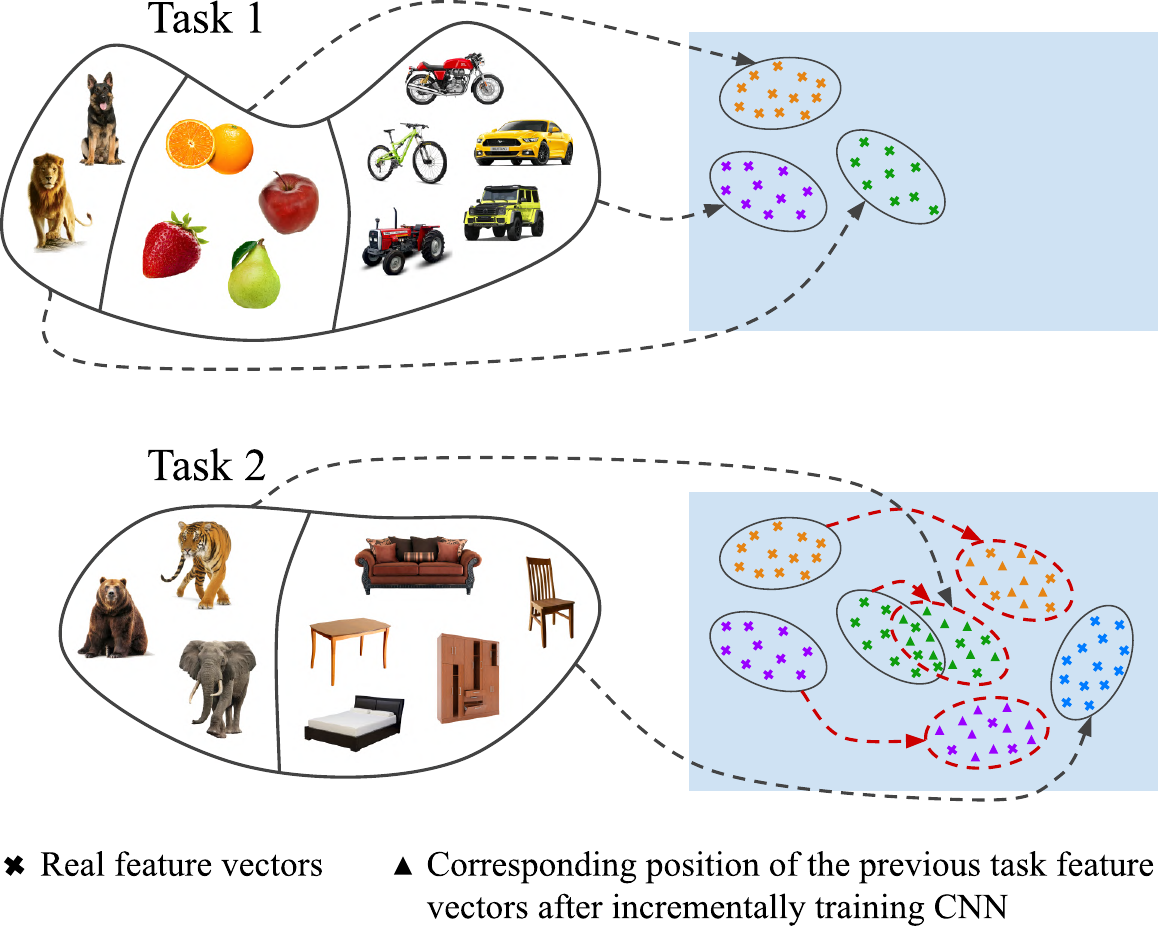}
\end{center}
   \caption{Feature transformation due to incremental training of the CNN: The top row indicates the feature vectors for each class in the feature space after learning Task 1. When the feature extractor (CNN) is trained on Task 2, the previous clusters become invalidated. The feature vectors in the Task 1 clusters need to be transformed to their corresponding new positions according to the current (updated) feature extractor.\vspace{-0.25cm}}
\label{fig:intro}
\end{figure}

Class Incremental Learning (CIL)~\cite{rebuffi2017icarl,hou2019learning,belouadah2019il2m,zhao2020maintaining} is a widely studied setting in IL research area, particularly in image classification problems.
CIL requires the learning of new classes for each task, where a task is defined as a group of classes.
There are certain rules that must be followed while defining a task.
(i) every task should consist of the same number of classes  (ii) there should not be any overlap between the classes in the tasks (i.e, if a class is present in task 1, it should not be present in any of the future tasks)  (iii) training samples are well-balanced across different classes.

However, this is not how humans and animals learn. They learn from imbalanced data, which is more realistic: each task does not necessarily consist of the same number of classes, tasks are intermixed in terms of classes, and class imbalance is prominent. In real scenarios, such as when a robot or a system is deployed into a real-world environment, the above three rules do not hold~\cite{wen2017fog}. This paradigm creates another branch of the problem, called Generalized Class Incremental Learning (GCIL)~\cite{mi2020generalized}, where each task is characterized by the following three quantities: the number of classes appearing in each task, the specific classes appearing in each task, and the number of samples per class. In the GCIL setting, these quantities are sampled from probabilistic distributions~\cite{mi2020generalized}. Hence, more diverse and realistic scenarios can be created by varying these distributions. This setup more closely resembles human learning.

Along with the challenge of catastrophic forgetting in previous CIL settings, GCIL has two other challenges, which are class imbalance and sample efficiency~\cite{mi2020generalized}. The Grow When Required (GWR)~\cite{marsland2002self} network, a Self-Organizing Map (SOM), learns a cluster by dynamically creating and removing nodes and edges when required. For each input feature vector, only the nodes in the neighborhood of that feature vector are trained according to a learning rule. This approach naturally aligns with class imbalance, sample efficiency, and the non-deterministic ordering of training samples in each task in the GCIL setting. A CNN acts as a feature extractor, creating a feature vector for each image input and then passing it to the GWR network to form clusters. To enhance the GWR's classification performance, the feature extractor should be capable of extracting class-specific features, so the feature extractor has to be trained incrementally. However, when the convolutional layers are adapted at each incremental step using a CIL method, the GWR nodes corresponding to prior knowledge may become nearly invalidated (~\cref{fig:intro}).

We have proposed the \emph{Self Augmenting Task Hallucinal Unified Representation (SATHUR)}, which re-initializes the GWR network at each incremental step, aligning it with the current feature extractor.
SATHUR takes the previous GWR nodes and exemplar feature vectors, which are extracted from exemplar samples by the current feature extractor, as input. It then trains a hallucinator to create a set of augmented feature vectors. The GWR network is subsequently re-initialized by training on these augmented feature vectors. The re-initialized GWR network is then trained on a mixture of feature vectors that are extracted from new task data and exemplar data. We have validated our approach using the CIFAR-100 and CORe50 datasets. The usage of our method, in conjunction with state-of-the-art replay-based CIL methods, improves accuracy by a significant margin. Specifically, it surpasses by $3.70\%$ on CIFAR-100 and by $2.88\%$ on CORe50.

\section{Related Work}
\label{sec:Related work}
\textbf{Class Incremental Learning.} In general, two groups of incremental training protocols are considered in the current CIL literature. The first is multi-epoch CIL, where new tasks, consisting of new classes or patterns, arrive incrementally, and only data in the current task is available for model training. During training, data of each task can be passed through multiple epochs. The second is online CIL. In this case, although training data still arrives sequentially, this setup only allows the model to be trained on each sample once~\cite{he2020incremental,hayes2020lifelong}. 

We focus only on the multi-epoch CIL setting and refer to it as CIL in the rest of the paper. The main problem of CIL is catastrophic forgetting, where learning a new task leads to degradation of performance related to previously learned tasks. There are three main approaches to mitigate catastrophic forgetting in CIL: (i) Regularization-based methods, (ii) Parameter-isolation-based methods (iii) Replay-based methods.

Regularization-based methods incorporate penalization terms into their objective functions to address discrepancies. These discrepancies typically exist between old and new models. This is often achieved by establishing comparisons across various elements. These elements include output logits~\cite{li2017learning, rebuffi2017icarl}, intermediate features~\cite{douillard2020podnet,
hou2019learning,
liu2023online,
simon2021learning}, and prediction heatmaps~\cite{deng2009imagenet}.

Parameter-isolation-based methods aim to increase the trainable model parameters at each incremental step, effectively counteracting the problem often seen with parameter overwriting. There are two primary strategies within this category. One approach involves gradually increasing the neural network's size to accommodate incoming data~\cite{huang2019neural,
rusu2016progressive,
wang2022foster,
xu2018reinforced,
yan2021dynamically}. The second strategy involves freezing a section of network parameters, ensuring that prior class knowledge remains preserved~\cite{abati2020conditional,
kirkpatrick2017overcoming,
liu2021adaptive,
zenke2017continual}. 

Replay-based methods are based on the idea that there's a specific, yet small, memory allowance for keeping a few old-class exemplars in memory compared to the new class data. These exemplars can be used to re-train the model in each new incremental step~\cite{douillard2020podnet,hou2019learning,liu2021rmm,liu2020mnemonics,rebuffi2017icarl,wang2022memory,wu2019large}. This re-training process usually has two parts: the first part involves training the model on all new class data and the small amount of old class exemplars, while the second part fine-tunes the model using a balanced subset with an equal number of samples from each class~\cite{douillard2020podnet, hou2019learning,liu2021adaptive,liu2021rmm,yan2021dynamically}.

\textbf{Generalized Class Incremental Learning.} GCIL was introduced by Fei Mi\textit{ et al.}~\cite{mi2020generalized}. They used a replay-based method~\cite{rebuffi2017icarl} in combination with mixup~\cite{zhang2017mixup}, a method they referred to as ReMix~\cite{mi2020generalized}, to address the challenges associated with GCIL. The herding technique was used to select exemplars from different classes during each incremental task training. Subsequently, mixup was applied to the mini-batches containing samples from both the current task and the exemplars. Mixup creates virtual training samples through the linear interpolation of raw training samples.

\textbf{Grow When Required network.} GWR~\cite{marsland2002self} network consists of nodes with weight vectors and connecting edges that establish node neighborhoods signifying similar perceptions. The system is dynamic, creating and removing nodes and edges based on the learning rule  detailed in ~\cref{sec: proposed solution}.

\section{Methods}
\label{sec:methods}
\subsection{Problem Definition}
\label{sec:problem definition}
As proposed in Mi \emph{et al.} \cite{mi2020generalized}, GCIL is formulated by structuring every task from a task-dependent distribution. We denote the complete set of available classes as $\mathcal{S}$ with size $n$. Given a sequence of tasks, $C_t$ is the set of samples that are present in the task $t$, and it is sampled from a task-dependent distribution $\mathcal{H}(t)$.
\begin{equation}
C_t\sim\mathcal{H}(t)
\label{eq:ct_ht}
\end{equation}
A probabilistic formulation of $\mathcal{H}(t)$ can be formed through three steps.
\begin{equation}
K_t\sim\mathcal{D}(t)
\label{eq:kt}
\end{equation}
The number of classes $K_t$ to appear in task $t$ follows a task-dependent discrete distribution $\mathcal{D}(t)$. Different scenarios regarding the number of appearing classes in each task can be simulated through different choices of $\mathcal{D}(t)$.
\begin{equation}
S_t\sim\mathcal{R}(W_t^1, K_t)
\label{eq:st}
\end{equation}
Classes appearing in task $t$ are modeled as a random vector $S_t \in {\rm I\!R}^n$. $S_t$ is a binary indicator vector with ones corresponding to classes appearing in $t$. $\mathcal{R}$ depends on the class number $K_t$ and a class appearance weight vector $W_t^1 \in {\rm I\!R}^n$. Each entry of $W_t^1$ represents the appearing probability of the class in task $t$. Classes with larger weights are more likely to appear in the task.
\begin{equation}
C_t\sim\mathcal{M}(W_t^2, S_t)
\label{eq:ct}
\end{equation}
To determine the sample size of each appearing class in $S_t$, which is encoded as random vector $C_t$. $C_t$ follows a distribution, which depends on the appearing class $S_t$ and a class sample size weight vector $W_t^2 \in {\rm I\!R}^n$. $W_t^2$ the sample size of each class appearing in task $t$, and it can model different degrees of class imbalance within a task. In~\cref{sec:implementation details}, we outline our choices for the distributions $\mathcal{D}(t)$, $W_t^1$ and $W_t^2$. Learning in the presence of this realistic presentation of classes is the GCIL problem.

\subsection{Proposed Solution}
\label{sec: proposed solution}
\begin{figure}[t]

\begin{subfigure}{.47\textwidth}
  \centering
  \includegraphics[width=\linewidth]{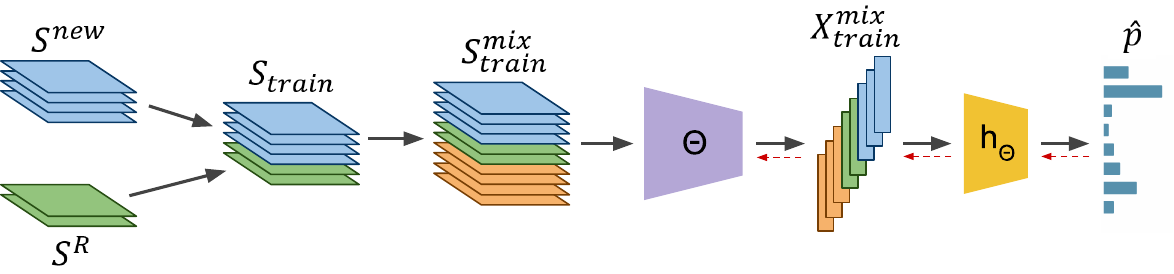}  
  \caption{Incremental training of feature extractor $\Theta$}
  \label{fig:sub-architecture01}
\end{subfigure}\hfill
\vspace{0.4cm}
\begin{subfigure}{.47\textwidth}
  \centering
  \includegraphics[width=\linewidth]{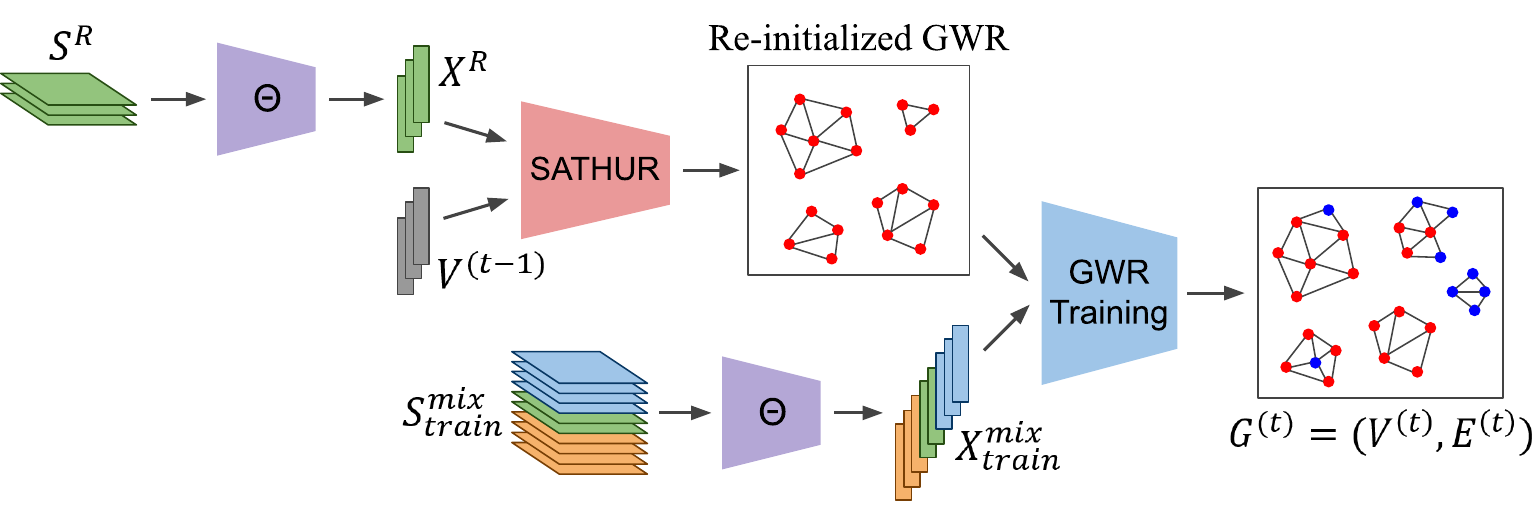}  
  \caption{Re-initializing GWR using SATHUR and training on new task data}
  \label{fig:sub-architecture02}
\end{subfigure}

\caption{GWR-based GCIL enabled by SATHUR.
(a) Training set $S^{train}$ incorporates new training samples $S^{new}$ and exemplars $S^R$. $S^{mix}_{train}$ is a balanced training set formed from $S_{train}$ by adding augmented samples generated through mixup. The feature extractor $\Theta$ is then trained using $S^{mix}_{train}$.
(b) Exemplar features $X_R$ are extracted from exemplar samples $S^R$ using updated $\Theta$ as the feature extractor. By using $X^R$ and previous task GWR nodes $V^{(t-1)}$ as the inputs to the SATHUR, GWR is re-initialized. Updated feature extractor $\Theta$ is used to extract the features $X^{mix}_{train}$ from $S^{mix}_{train}$. Re-initialized GWR is trained on $X^{mix}_{train}$ to learn the unified representation. Red nodes represent re-initiated nodes before training on new task data, and blues nodes represent the nodes that are created during new task training.\vspace{-0.25cm}}
\label{fig:architecture}
\end{figure}

We propose SATHUR, an effective method that can incrementally re-initialize the GWR network according to the updated feature extractor and train GWR on the new task. Our method can be used as a plugin method along with existing replay-based CIL methods.\\

\noindent \textbf{Incremental training of feature extractor}. 
At each incremental training step, new task data is combined with the exemplars to form a unified training set.
The mixup~\cite{zhang2017mixup} is applied to the unified training set to generate new augmented samples.
The augmented samples are generated by taking convex combinations of pairs of inputs and their labels, which create more diverse samples for underrepresented classes by mixing them with over-represented ones.
An augmented training sample $(\bar{x}, \bar{y})$ is generated by linear interpolation between raw training samples $(x_i, y_i)$ and $(x_j, y_j)$ as shown in~\cref{eq:remix_interpolation}.
This increase in samples assists the model in learning better representations of under-represented classes, while facilitating the learning of smooth decision boundaries between all classes.
\begin{equation}
  \bar{x}=\lambda x_i+(1-\lambda)x_j, ~\bar{y}=\lambda y_i+(1-\lambda)y_j
  \label{eq:remix_interpolation}
\end{equation}
\noindent where $\lambda\sim Beta(\alpha, \alpha)$, with hyperparameter $\alpha \in (0, \infty)$\vspace{0.25cm}

As illustrated in~\cref{fig:sub-architecture01}, we combine the samples corresponding to the new task $S^{new}$ and exemplars $S^R$ to create a unified training set $S_{train}$. We apply mixup on $S_{train}$ to create balanced training set $S^{mix}_{train}$, which is then used to train the feature extractor, $\Theta$. After training, $\Theta$ is able to extract the important high-level features corresponding to $S^{mix}_{train}$.

\vspace{0.2cm}
\noindent \textbf{GWR-based GCIL}. 
GWR networks are suitable for learning imbalanced feature distributions in the GCIL setting due to their ability to introduce new neurons when it is required to match new inputs.

GWR~\cite{marsland2002self} network is initialized by randomly selecting two feature vectors as inputs. The weights ($w$) and labels ($l$) of the nodes correspond to those of the feature vectors.
The best matching unit ($s$) is the node that most closely resembles the input. For each input, an edge connection is generated between the best matching unit and the second-best matching unit ($t$).
The GWR network adds new nodes based on two conditions. First, the activity ($a$) of the best matching node, which is inversely related to the Euclidean distance from the input, should fall below a certain activity threshold ($a_T$). This indicates that the current best matching node does not closely resemble the input. 
Secondly, the node's firing counter ($h$), must also fall below a certain firing threshold ($h_T$). 
\begin{equation}
  h_s(t) = h_0 - \frac{1}{\alpha_b}(1 - e^{-\alpha_b t/\tau_b})\\
  \label{eq:bmu_firing_counter}
\end{equation}
\begin{equation}
  h_n(t) = h_0 - \frac{1}{\alpha_n}(1 - e^{-\alpha_n t/\tau_n})\\
  \label{eq:neighbour_firing_counter}
\end{equation}
\noindent where $h_0$ is the initial firing counter value. $\alpha_b, \alpha_n$ and $\tau_b, \tau_n$ are the constants controlling behaviour of the curve.\vspace{0.25cm}
Whenever the criterion for adding a new node is not met, the weights of the best matching node and its directly connected neighboring nodes are adjusted to learn the new input. $\epsilon_b$ and $\epsilon_n$ are learning rates for the best matching node and neighboring nodes, respectively.
The firing counter of the best matching node is updated using~\cref{eq:bmu_firing_counter}.
The firing counters of directly connected neighboring nodes to the best matching node are updated using~\cref{eq:neighbour_firing_counter}.
The edge connections have an associated \emph{age}, which is initially set to zero and is incremented at each time step for each edge connected to the best matching node. The only exception is the edge that links the best matching and second-best matching units, whose \emph{age} is reset to zero. Edges that exceed a certain age threshold, $age_{max}$, are removed. Any node without direct neighbors, i.e., without edge connections, is removed.

During the inference phase of GWR, the nearest neighbors to the input feature vector are determined based on Euclidean distance. The label of the input feature vector is then predicted as the most common label among these nearest neighbors.

\begin{table}
  \centering
    \begin{tabular}{lp{7.3cm}}
    \toprule
     \multicolumn{2}{l}{\textbf{Algorithm 1:} GWR training}\\
    \midrule
    1 & Initialize the network by randomly selecting two inputs to form nodes with corresponding labels.\\
    2 & \textbf{for} each input feature $x_i$:\\
    3 & \hspace{3mm}\textbf{for} each node $i$ in the network: calculate $\lVert x_i - w_i \rVert$\\
    4 & \hspace{3mm}Select nodes $s, t$; $s= argmin_{v \in \mathcal{V}}{\{\lVert x_i - w_i \rVert\}}$\\
      & \hspace{3mm}and $t= argmin_{v \in \mathcal{V} \setminus \{s\}}{\{\lVert x_i - w_i \rVert\}}$\\
    5 & \hspace{3mm}\textbf{if} $(s,t)$ not in $\mathcal{E}$: $\mathcal{E} = \mathcal{E} \cup \{(s,t)\}$\\
    5 & \hspace{3mm}\textbf{else}: $age_{(s,t)} = 0$\\
    6 & \hspace{3mm}Calculate activity for $s$; $a_s = exp(-\lVert x_i - w_s \rVert)$\\
    7 & \hspace{3mm}\textbf{if} $a_s<a_T$ and $h_s<h_T$:\\
    8 & \hspace{6mm}Add new node $r$; $w_r = (w_s+x_i)/2$ and $l_r = l_i$\\
    9 & \hspace{6mm}$\mathcal{E}=\mathcal{E}\cup\{(r,s), (r,t)\}$\\
    10 & \hspace{6mm}$\mathcal{E}=\mathcal{E}\setminus\{(s,t)\}$\\
    11 & \hspace{3mm}\textbf{else}:\\
    12 & \hspace{6mm}Adapt the positions of the $s$ and its neighbors;\\
       & \hspace{6mm}$\Delta w_s = \epsilon_b \times h_s \times (x_i - w_s)$\\
       & \hspace{6mm}$\Delta w_j = \epsilon_n \times h_j \times (x_i - w_j)$\\
    13 & \hspace{6mm}\textbf{for} all edges ending at $s$: $age_{(s,q)} = age_{(s,q)} + 1$\\
    14 & \hspace{6mm}Update the $h_s$ using~\cref{eq:bmu_firing_counter}\\
    15 & \hspace{6mm}Update the $h_j$ of the neighbors of $s$ using~\cref{eq:neighbour_firing_counter}\\
    16 & \hspace{3mm}\textbf{for} all $(p,q) \in \mathcal{E}$:\\
    17 & \hspace{6mm}\textbf{if} $age_{(p,q)}>age_{max}$: $\mathcal{E}=\mathcal{E}\setminus\{(p,q)\}$\\
    18 & \hspace{3mm}\textbf{for} all $v \in \mathcal{V}$:\\
    19 & \hspace{6mm}\textbf{if} no edge end at $v$: $\mathcal{V} = \mathcal{V} \setminus \{v\}$\\
    \bottomrule
    \end{tabular}
    \label{tab:gwr training}
\end{table}
\begin{table}
  \centering
    \begin{tabular}{lp{7.3cm}}
    \toprule
     \multicolumn{2}{l}{\textbf{Algorithm 2:} SATHUR (task $t, t \geq 2$)}\\
    \midrule
    \multicolumn{2}{l}{\textbf{Input\hspace{0.26cm}:} GWR nodes $V^{(t-1)}$ of task $(t-1)$}\\
    \multicolumn{2}{l}{\hspace{1.235cm} Exemplar features $X^R$ extracted by updated $\Theta$}\\
    \multicolumn{2}{l}{\textbf{Output:} Re-initialized GWR,}\\
    \multicolumn{2}{l}{\hspace{1.235cm} $G^{(t-1)*} = (V^{(t-1)*}, E^{(t-1)*})$}\\
    1 & \textbf{for} $epochs$ \textbf{do}:\\
    2 & \hspace{3mm}$V^{(t-1)}_j \gets$ choose $m$ samples from $V^{(t-1)}$\\
    3 & \hspace{3mm}\textbf{for} every $V^{(t-1)}_{j,k} \in V^{(t-1)}_j$:\\
    4 & \hspace{6mm}Select $X^R_{j,k}$ from $X^R$; same class as $V^{(t-1)}_{j,k}$\\
    5 & \hspace{3mm}$X^R_j \gets$ all selected exemplar samples $X^R_{j,k}$\\
    6 & \hspace{3mm}$X^{aug} = P^2(P^1(V^{(t-1)}_j) + X^R_j)$; $P = P^1 \cup P^2$\\
    8 & \hspace{3mm}$X^{aug}_{train} = X^R \cup X^{aug}$\\
    9 & \hspace{3mm}Train hallucinator $P$, using classifier $h_P$\\
    10 & Using updated hallucinator $P$, create $X^{aug}_{train}$\\
    11 & Train a GWR network on $X^{aug}_{train}$ using Algorithm 1\\
    \bottomrule
    \end{tabular}
    \label{tab:sathur_training}
\end{table}

\begin{figure}[t]
  \centering
  \includegraphics[width=0.475\textwidth]{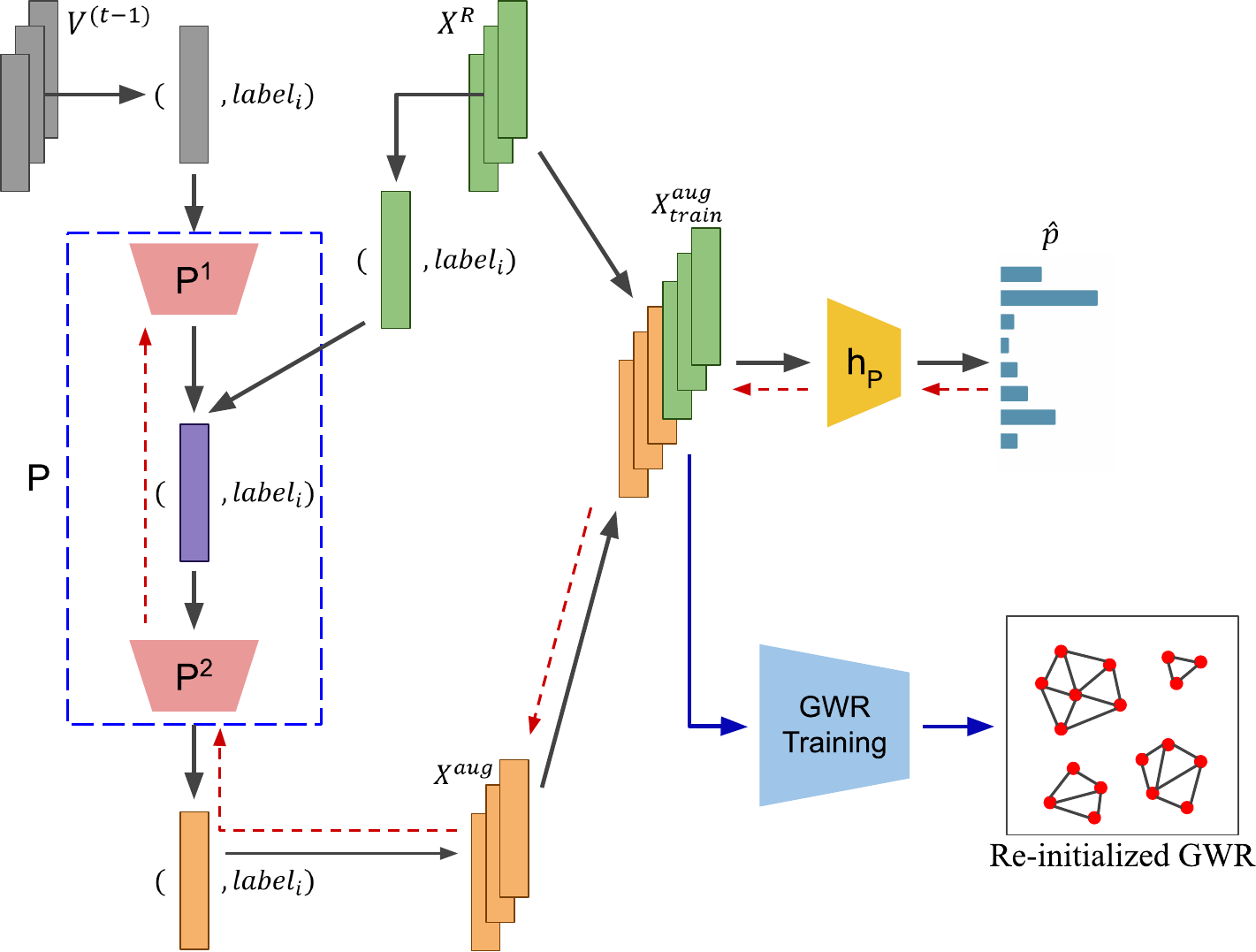}
  \caption{
  Re-initializing GWR using SATHUR: 
  Following the incremental training of feature extractor $\Theta$, previous task GWR nodes $V^{(t-1)}$ and exemplar features $X^R$ from the same class are sampled.
  Sampled previous task GWR nodes are passed into $P^1$ and the output is combined with the corresponding class exemplar features and passed into $P^2$ to obtain augmented features $X^{aug}$. 
  A unified training set $X^{aug}_{train}$ is created by adding $X^{aug}$ to $X^R$. The hallucinator $P$ is trained end-to-end along with the classification algorithm $h_P$. Dotted red arrows indicate the flow of gradients during back-propagation. Once the hallucinator $P$ is trained, then a unified training set $X^{aug}_{train}$ is generated by combining $X^R$ and augmented features $X^{aug}$, created by the trained hallucinator. GWR is re-initiated by training $X^{aug}_{train}$ on the GWR algorithm. As a result, new nodes are created for previous classes validating the new task feature extractor.\vspace{-0.25cm}}
  \label{fig:sathur}
\end{figure}

\vspace{2mm}
\noindent \textbf{Re-initializing GWR using SATHUR}.
SATHUR is used to re-initialize the previous task GWR network $G^{(t-1)}$, adapting to the updated feature extractor $\Theta$.
As shown in~\cref{fig:sathur}, at every training iteration we choose $m$ samples from previous task GWR nodes $V^{(t-1)}$ and form a node batch. Exemplar features $X^R$ are extracted using updated feature extractor $\Theta$.
We form an exemplar batch $X^R_j$ by choosing samples from $X^R$, such that for every node in the node batch, there is a corresponding exemplar feature that belongs to the same class.
The intermediate output obtained by passing $V^{(t-1)}$ to $P^1$ is combined with corresponding samples in $X^R$ and passed to $P^2$ to generate augmented features $X^{aug}$.
Each augmented example is of the form $(x',y)$, where $x'=P(x,v;\textbf{w}_P)$. Here, $(x,y)$ is a sample from $X^R$, and $(v,y)$ is the corresponding sample from $V^{(t-1)}$. $\textbf{w}_P$ represents the parameters of the hallucinator, $P$. 
An augmented training set, $X^{aug}_{train}$, is formed by adding the set of augmented features, $X^{aug}$, to the set of exemplar features, $X^R$.
The hallucinator $P$ is trained end-to-end along with the classification algorithm $h_P$.
After training the hallucinator $P$, an augmented training set $X^{aug}_{train}$ is created. $X^{aug}_{train}$. 
The GWR network is re-initialized by training with $X^{aug}_{train}$ using Algorithm 1.

\section{Experiments}
\label{sec:experiments}
\subsection{Baseline Methods}
\label{sec:baseline methods}

\noindent\textbf{ReMix}~\cite{mi2020generalized}: Mixup is applied to the new task data and previously stored exemplars to generate more augmented data. The model is jointly trained on new task data, exemplar data, and augmented data.

\noindent\textbf{Mnemonics}~\cite{liu2020mnemonics}: A bilevel optimization framework was used to distill new class data into exemplars before discarding them. The aim of this method is to improve the quality of exemplars without inflating their number.

\noindent\textbf{MRDC}~\cite{wang2022memory}: The aim of this method is to establish a balance between the quality and quantity of exemplars. This was accomplished through image compression, utilizing the JPEG algorithm, which resulted in each exemplar being uniformly downsampled.

\noindent\textbf{GWR}~\cite{marsland2002self}: Pre-trained convolutional layers are used as the feature extractor and GWR network is trained only on new task data.

\noindent\textbf{Full}: At each task, the model is trained on all the task data that have been arrived at the model. This is a common performance upper bound in CIL.

To ensure a fair comparison between the methods, we used ImageNet pre-trained 32-layer ResNet~\cite{he2016deep} as the base initialization for the feature extractor $\Theta$. For ReMix, Mnemonics, and MRDC, $\Theta$ is incrementally trained using the respective algorithms. In the case of Full, at every task, base initialization is done, and then it is trained on all the data that the model has encountered up to that particular task. For GWR, feature extractor $\Theta$ cannot be trained incrementally without SATHUR. Therefore, we used the ImageNet pre-trained 32-layer ResNet~\cite{he2016deep} as the feature extractor $\Theta$ without training it incrementally. 

\subsection{Datasets}
\label{sec:datasets}
We compare the model's performance on CIFAR-100~\cite{krizhevsky2009learning} and CORe50~\cite{lomonaco2017core50} datasets.

\noindent\textbf{CIFAR-100:} Contains 60,000 RGB images, each sized 32$\times$32 pixels, spread across 100 classes. Each class contains 500 training images and 100 testing images. We train 20 tasks incrementally, each containing 1,000 images. In each experiment, 20,000 images are randomly selected from the collection of 60,000 images for training.

\noindent\textbf{CORe50:} Contains short 15-second video sequences of an object moving. It has 10 object categories, each with 5 distinct domestic objects, recorded under 11 different environmental conditions (e.g., various backgrounds, various illuminations, outdoors/ indoors, etc.), 8 for training, and 3 for testing. The videos were originally recorded at 20 fps, but we sample them at 2 fps. Due to the smoothness of the videos, down-sampling the video frame rate is a common practice with CORe50~\cite{hayes2019memory, hayes2020lifelong}.
The training set contains 12,000 RGB images, and the testing set contains 4,500 images, each sized 128$\times$128 pixels, spread across 10 classes. We incrementally train 10 tasks with 600 images in each task. In each experiment, 6,000 images are randomly selected from the collection of 12,000 images for training.

\subsection{Implementation Details}
\label{sec:implementation details}

\begin{figure*}[t]
\begin{center}
 	\centering
    \subfloat[]{\includegraphics[width=0.32\linewidth]{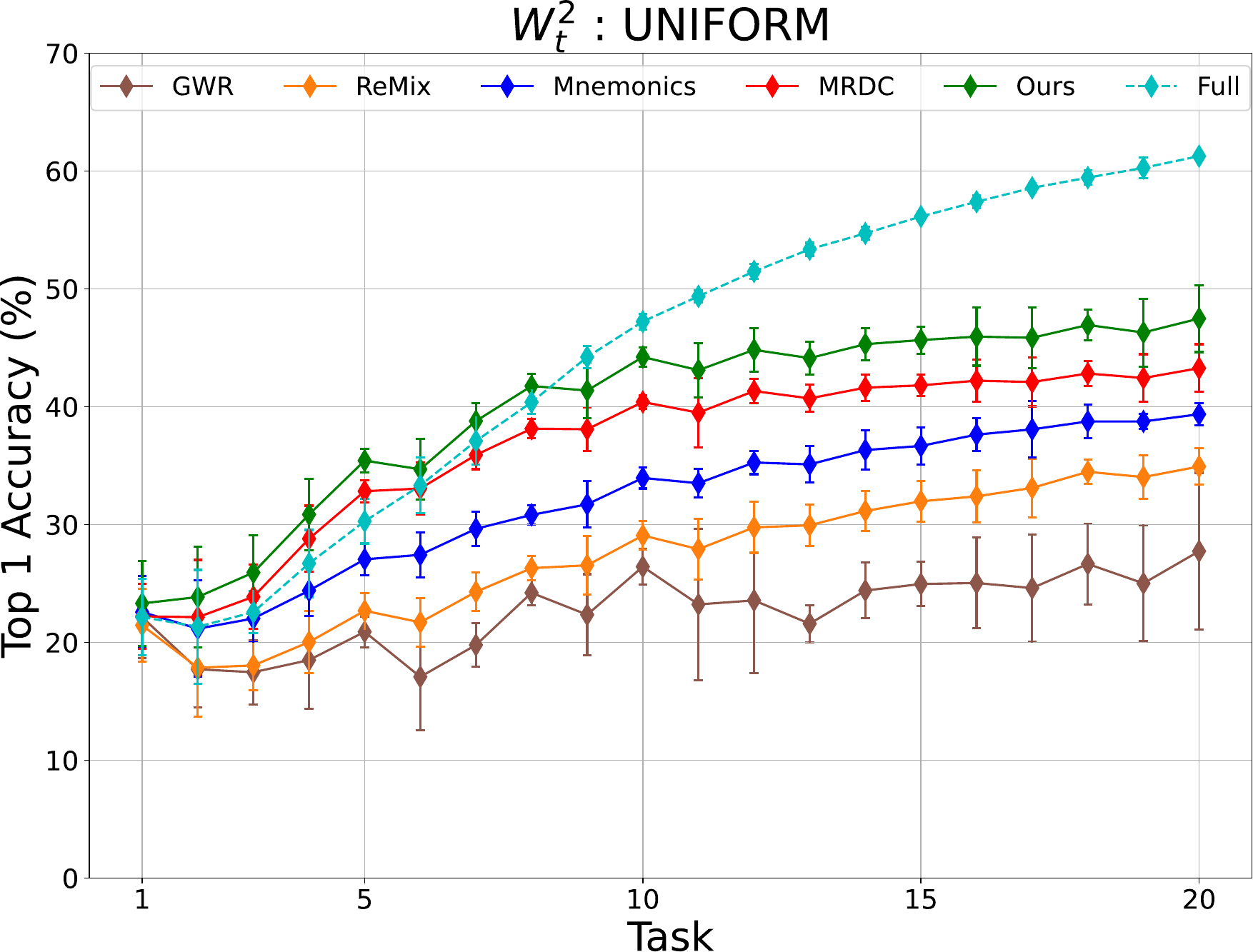}\label{subfig:cifar100_g1}}
	\hfil
    \subfloat[]{\includegraphics[width=0.32\linewidth]{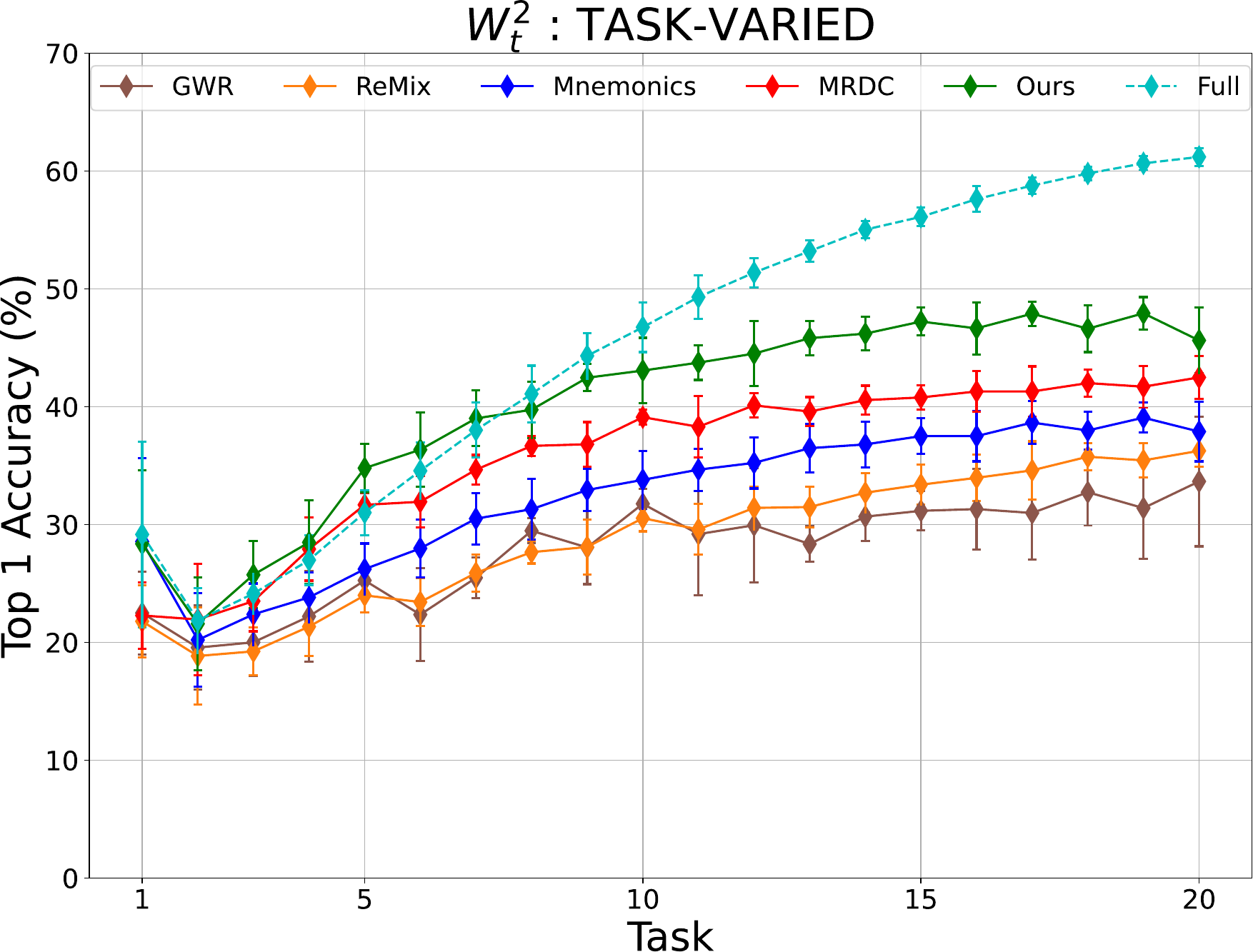}\label{subfig:cifar100_g2}}
    \hfil	
    \subfloat[]{\includegraphics[width=0.32\linewidth]{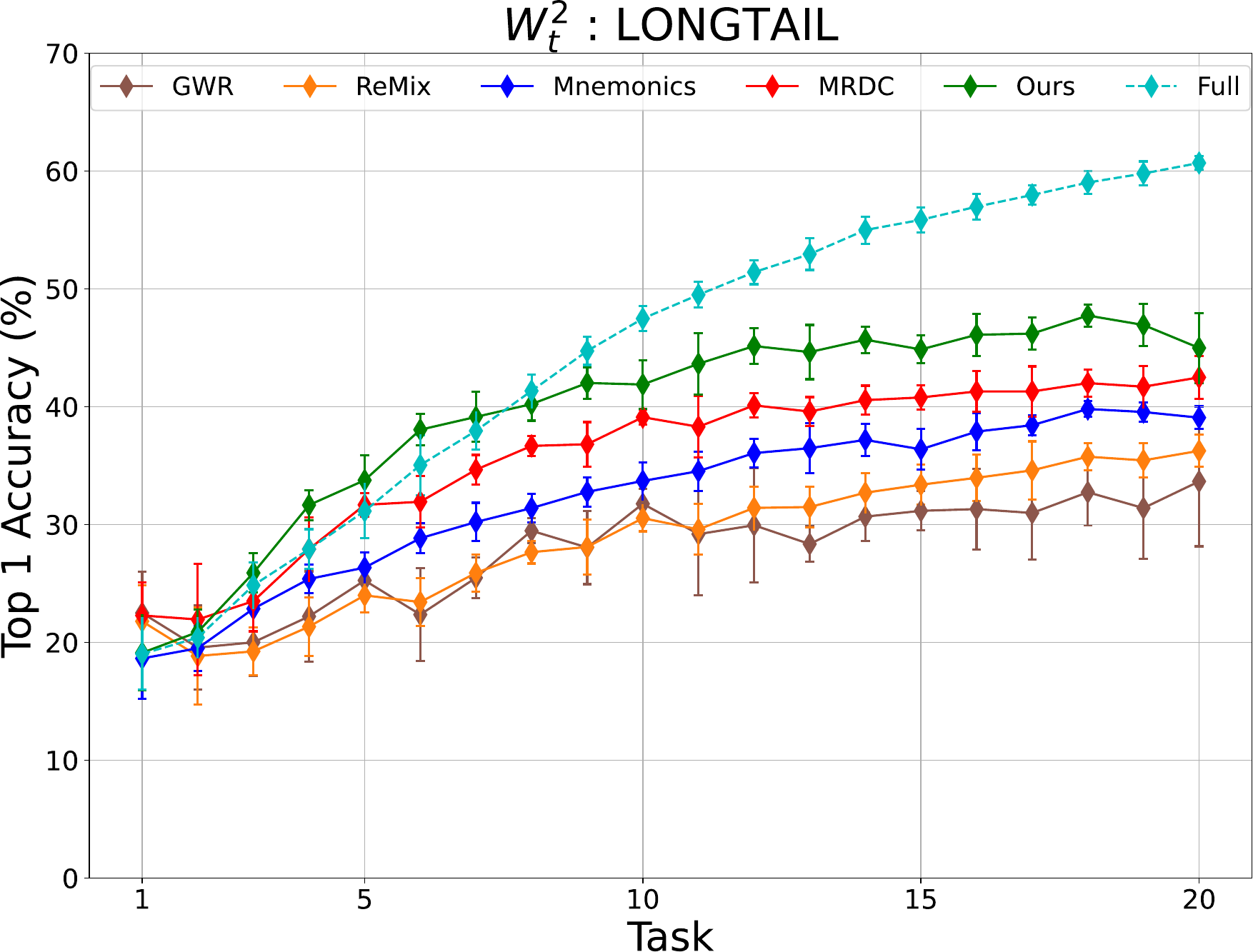}\label{subfig:cifar100_g3}}
    \hfil
\end{center}
\vspace{-0.4cm}
\caption{\small
Performances with varying $W_t^2$ on CIFAR-100: At each incremental training task, mean top-1 accuracy and standard deviation averaged over 5 runs with different random seeds are plotted. Our method significantly outperforms existing methods.
}
\label{fig:cifar100 results}
\vspace{-0.4cm}
\end{figure*}
\begin{figure*}[t]
\vspace{0.4cm}
\begin{center}
 	\centering
    \subfloat[]{\includegraphics[width=0.32\linewidth]{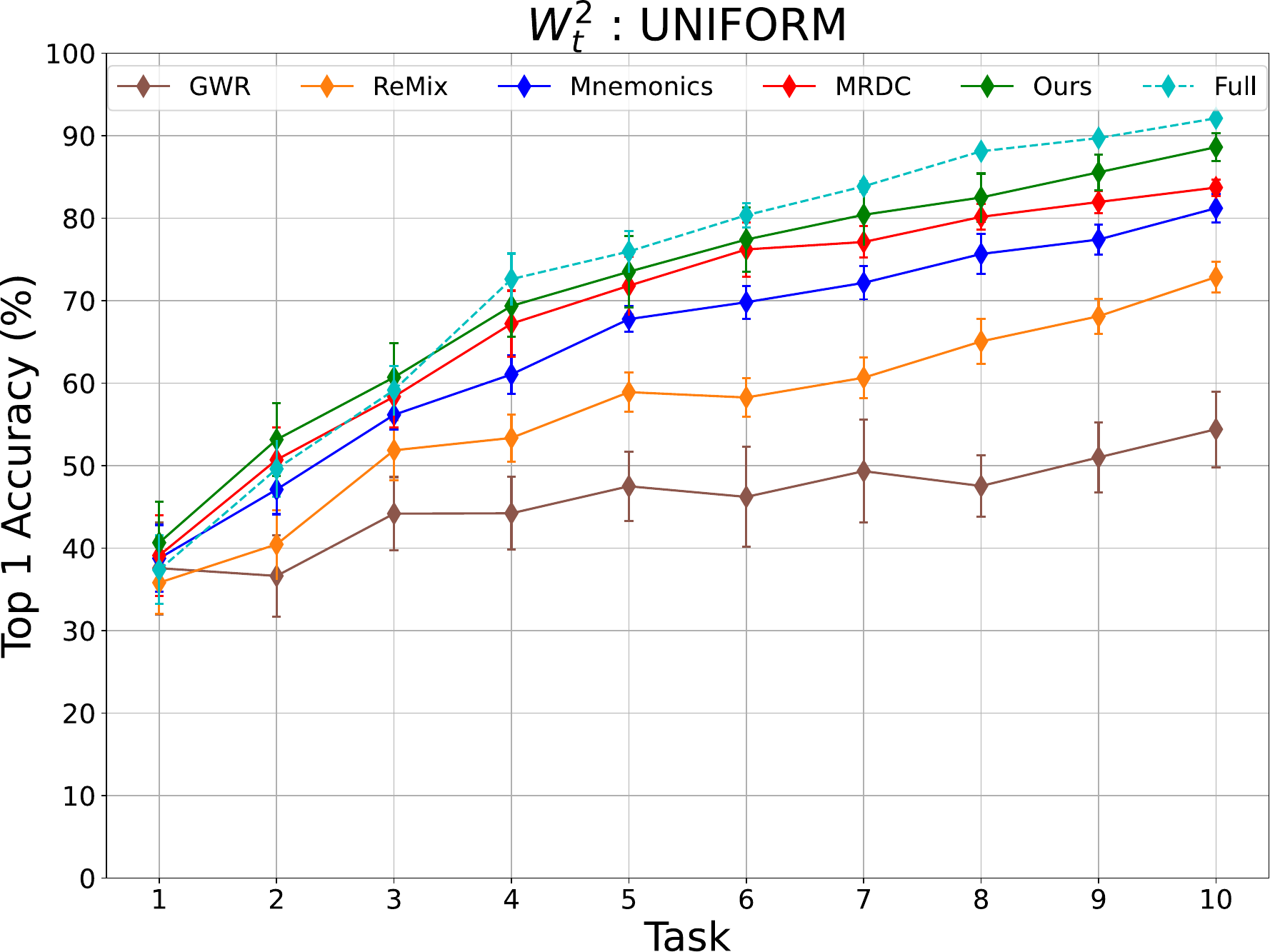}\label{subfig:core50_g1}}
    \hfil
    \subfloat[]{\includegraphics[width=0.32\linewidth]{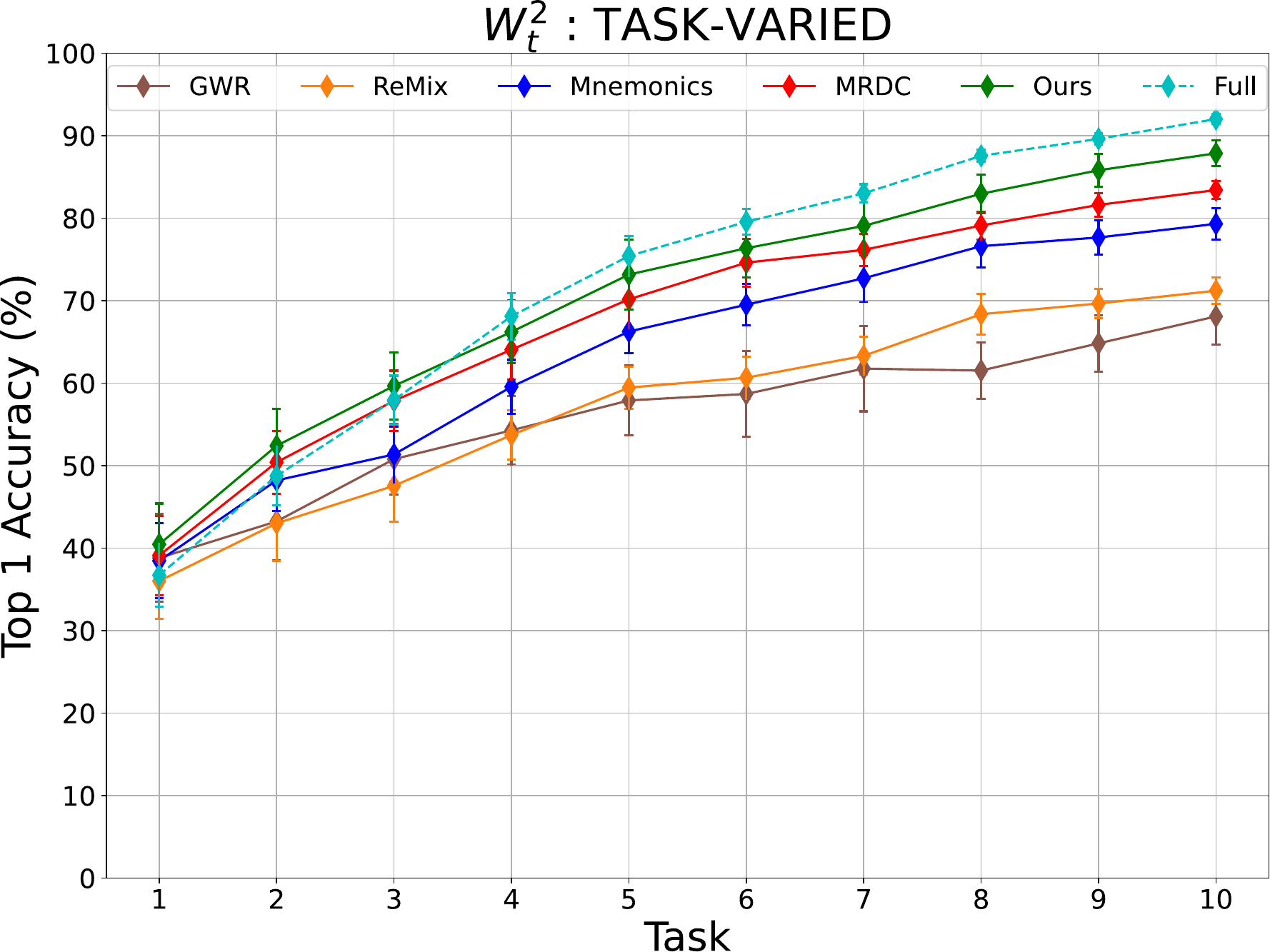}\label{subfig:core50_g2}}
    \hfil
    \subfloat[]{\includegraphics[width=0.32\linewidth]{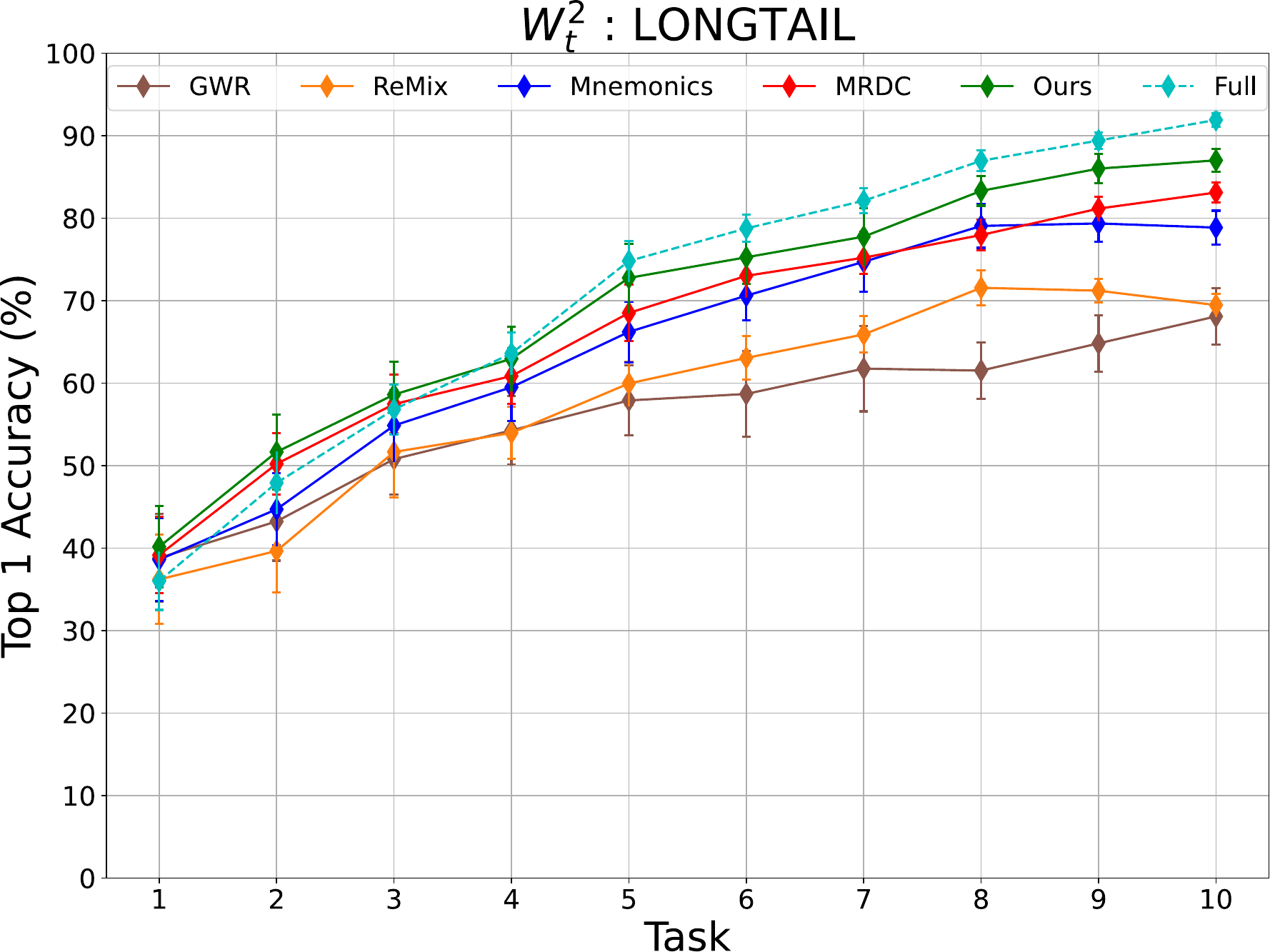}\label{subfig:core50_g3}}
    \hfil
\end{center}
\vspace{-0.25cm}
\caption{\small 
Performances with varying $W_t^2$ on CORe50: At each incremental training task, mean top-1 accuracy and standard deviation averaged over 5 runs with different random seeds are plotted. Ours method significantly outperforms existing methods and reaches very close to the upper limit ``Full''.
}
\label{fig:core50 results}
\vspace{-0.25cm}
\end{figure*}
\noindent\textbf{Feature extractor $\Theta$.} At each incremental task training, a 32-layer ResNet~\cite{he2016deep} is used as feature extractor $\Theta$ and it is trained by stochastic gradient descent with 60 epochs. $\alpha$ of ReMix is set to 1.2. The learning rate starts from 0.1 and is divided by 10 after 40 and 50 epochs; weight decay is 1e-3 and momentum is 0.9. 

\vspace{2mm}
\noindent\textbf{SATHUR.} For our hallucinator $P$, we use a two layer MLP for $P^1$ and three layer MLP for $P^2$ with ReLU as the activation function. $P$ is trained by stochastic gradient descent with 100 epochs.  The learning rate starts from 0.1 and is divided by 10 after 60 and 80 epochs; weight decay is 1e-3 and momentum is 0.9. For the GWR network activity threshold and firing threshold are set to 0.65 and 0.11 respectively.

We evaluate our method on benchmark as done in ReMix~\cite{mi2020generalized}. For CIFAR-100 dataset, we set $\mathcal{D}(t)$ as a uniform distribution $\mathcal{U}(1, 100)$, and $W^1_t$ as a uniform distribution over all classes. For CORe50 dataset, we set $\mathcal{D}(t)$ as a uniform distribution $\mathcal{U}(1, 10)$, and $W^1_t$ as a uniform distribution over all classes. Three variations of $W^2_t$ are tested;

\noindent\textbf{UNIFORM:} $W^2_t$ is a fixed uniform distribution over all classes in $S$.

\noindent\textbf{TASK-VARIED:} $W^2_t$ varies across different tasks by adding independent Gaussian noises (0 mean and $20\%$ of uniform class weight as standard deviation) to each class weight of UNIFORM.

\noindent\textbf{LONGTAIL:} $W^2_t$ is a fixed long-tailed distribution. The weight $W^2_{t,i}$ for class i in the long-tailed distribution is generated by an exponential function $W^2_{t,i} = \mu^i$~\cite{cui2019class}. Different $\mu$’s correspond to different degrees of class imbalance. In our setting, the largest weight is 5 times larger than the smallest.

Models are evaluated by TOP-1 accuracy on the balanced test set consisting of all classes that appeared so far.


\section{Results}
\label{sec:results}

~\Cref{fig:cifar100 results} and \Cref{fig:core50 results} show the performance of the methods, with the CIFAR-100 and CORe50 datasets respectively, measured by mean top-1 accuracy and standard deviation, based on five experimental runs each.
We compare (see~\cref{sec:baseline methods}) ReMix, Mnemonics, MRDC, GWR, and Full with our method. Our method outperforms the state-of-the-art methods by significant margins in different GCIL setups.
When $W^2_t$ = LONGTAIL, it surpasses the state-of-the-art method, MRDC, on CIFAR-100 and CORe50 by $3.70\%$ and $2.88\%$, respectively.
During early incremental tasks, our method performs better than ``Full''. This is because SATHUR is creating augmented features that are so close to real features.

\subsection{Ablation study}
\label{sec:ablation study}

\begin{table}
  \centering
    \begin{tabular}{@{}lcr@{}}
    \toprule
    Method & CIFAR-100 & CORe50 \\
    \midrule

    GWR & 27.80 & 55.98\\
    ReMix & 28.76 & 58.26\\
    Mnemonics & 32.25 & 64.64\\
    MRDC & 35.72 & 66.66\\
    Ours (MRDC+GWR+SATHUR) & \textbf{39.42} & \textbf{69.54}\\
    \bottomrule
    \end{tabular}
    \caption{Top-1 accuracy averaged over 20 tasks for CIFAR-100 and 10 tasks for CORe50 when $W^2_t$ = LONGTAIL. The contribution from SATHUR is essential for the good results achieved.\vspace{-0.25cm}}
    \label{tab:ablation study}
\end{table}

In \cref{tab:ablation study}, our approach is evaluated against GWR and MRDC.
The fact that training GWR using the features extracted from the pretrained $\Theta$ as the feature extractor fails significantly shows that $\Theta$ needs to be optimally trained at each incremental task.
MRDC exhibits good performance in the GCIL setting compared to Mnemonics and ReMix, by maintaining a higher number of compressed training samples within the memory buffer.
Therefore, we used MRDC in conjunction with mixup to train $\Theta$ at each incremental task.
Hence, the use of SATHUR is crucial to transform the previously generated nodes across incremental tasks.
Our method outperformed GWR by a large margin of $11.62\%$ on CIFAR-100 and $13.56\%$ on CORe50 datasets.
Hence, the use of SATHUR is crucial to transform the previously generated nodes across incremental tasks.
Our method outperformed GWR by a large margin of $11.62\%$ on the CIFAR-100 and $13.56\%$ on the CORe50 datasets.

By splitting the allocated memory between exemplar samples and GWR nodes, our method can be used as a plugin method with replay-based CIL methods.
Future research can be conducted in the area of adaptive memory optimization between exemplars and GWR node at each incremental step.

As the GWR is adapted only to the local neighborhood that is most similar to the input, catastrophic interference to parameters unrelated to the current input is prevented. This local adaptation is lightweight in terms of computation and memory requirements, compared to backpropagation-based learning with entire model adaptation. Once the feature extractor is trained, GWR can incrementally learn from even a very small amount of samples without fine-tuning the feature extractor. This makes our approach capable of performing collaborative cloud and edge computing~\cite{ren2019collaborative}.

\section{Conclusion}
\label{sec:conclusion}
Generalized Class Incremental Learning (GCIL) represents a more realistic continual learning paradigm where each incremental task is customized based on probabilistic distributions. Therefore, different realistic scenarios can be simulated by altering these distributions. GWR networks inherently possess the ability to learn distributions by managing class imbalance, sample efficiency, and the non-deterministic ordering of training samples. However, the performance of GWR is constrained by the fixed feature extractor used to extract feature vectors from images. In this work, we propose a method to train the GWR network while incrementally adapting the feature extractor. We introduce a \emph{Self Augmenting Task Hallucination Unified Representation (SATHUR)} to re-initialize the GWR network at each incremental step, thereby adapting to the updated feature extractor. Our method's effectiveness in addressing the GCIL problem is demonstrated by our successful results with the CIFAR-100 and CORe50 datasets. In the context of continual learning, our research expands the capacity to learn from real-world data samples. It facilitates the accumulation of new knowledge while concurrently mitigating the issues of catastrophic forgetting and class imbalance.


{\small
\bibliographystyle{ieee_fullname}
\bibliography{bib}
}

\end{document}